\documentclass[letterpaper, 10 pt, conference]{ieeeconf}  

\IEEEoverridecommandlockouts                              

\usepackage{graphics} 
\usepackage{epsfig} 
\usepackage{times} 
\usepackage{amsmath} 
\usepackage{amssymb}  
\usepackage[caption=false]{subfig}
\let\labelindent\relax
\usepackage[shortlabels]{enumitem}
\usepackage{multirow}
\usepackage{booktabs}
\usepackage[nocomma]{optidef}

\newcommand\numberthis{\addtocounter{equation}{1}\tag{\theequation}}
\newcommand{\deltat}{\Delta t}
\newcommand{\jacembedsensing}{\mathbf{J}_{z}}
\newcommand{\jacaction}{\mathbf{J}_{a}}
\newcommand{\jacprevembedsensing}{\mathbf{J}_{z_{t-1}}}
\newcommand{\jacprevaction}{\mathbf{J}_{a_{t-1}}}
\newcommand{\prevlinearizednonlinlatentdynA}{\mathbf{\bar{A}}_{t-1}}
\newcommand{\prevlinearizednonlinlatentdynB}{\mathbf{\bar{B}}_{t-1}}
\newcommand{\prevlinearizednonlinlatentdync}{\mathbf{\bar{c}}_{t-1}}
\newcommand{\sensing}{\mathbf{s}}
\newcommand{\embedsensing}{\mathbf{z}}
\newcommand{\embedfunc}{\mathbf{f}_{enc}}
\newcommand{\inverseembedfunc}{\mathbf{f}_{dec}}
\newcommand{\discretelatentdynfunc}{\mathbf{f}_{dfd}}
\newcommand{\latentdynfunc}{\mathbf{f}_{fd}}
\newcommand{\invdynfunc}{\mathbf{f}_{id}}
\newcommand{\dynmodelnnparams}{\mathbf{\theta}_{d}}
\newcommand{\action}{\mathbf{a}}
\newcommand{\norm}[1]{\left\lVert#1\right\rVert}

\newcommand{\cartesianvelocity}{\dot{\mathbf{x}}}

\newcommand{\linearlatentdynA}{\mathbf{A}}
\newcommand{\linearlatentdynB}{\mathbf{B}}
\newcommand{\linearlatentdync}{\mathbf{c}}
\newcommand{\lagrangemultiplier}{\mathbf{\lambda}}
\newcommand{\eye}{\mathbf{I}}

\newcommand{\argmin}{\operatorname*{arg\:min}}

\newcommand{\T}{^{\textrm T}} 

\usepackage[bookmarks=true]{hyperref}

\usepackage[disable]{todonotes}
\newcommand{\GS}[1]{\todo[inline,color=blue!40]{G: #1}}

\newcommand{\YC}[1]{\todo[inline,color=red!40]{Y: #1}}

\selectfont
\title{\LARGE \bf Learning Latent Space Dynamics for Tactile Servoing}
\author{Giovanni Sutanto$^{1,2,3}$, Nathan Ratliff$^{1}$, Balakumar Sundaralingam$^{1,4}$, Yevgen Chebotar$^{1,3}$,\\
Zhe Su$^{2,3}$, Ankur Handa$^{1}$ 
and Dieter Fox$^{1,5}$
  \thanks{$^{1}$NVIDIA, USA.}%
  \thanks{$^{2}$Autonomous Motion Department, MPI-IS, T\"ubingen, Germany.}%
  \thanks{$^{3}$University of Southern California, Los Angeles, USA.}%
  \thanks{$^{4}$University of Utah, Salt Lake City, USA.}%
  \thanks{$^{5}$University of Washington, Seattle, WA, USA.}%
  \thanks{This research was supported in part by NVIDIA Research, National Science
    Foundation grants IIS-1205249, IIS-1017134, EECS-0926052, the
    Office of Naval Research, the Okawa Foundation, and the
    Max-Planck-Society.}
}

\begin{document}

\maketitle
\thispagestyle{empty}
\pagestyle{empty}

\begin{abstract}
	To achieve a dexterous robotic manipulation, we need to endow our robot with tactile feedback capability, \textit{i.e.} the ability to drive action based on tactile sensing. In this paper, we specifically address the challenge of tactile servoing, \textit{i.e.} given the current tactile sensing and a target/goal tactile sensing --- memorized from a successful task execution in the past --- what is the action that will bring the current tactile sensing to move closer towards the target tactile sensing at the next time step. We develop a data-driven approach to acquire a dynamics model for tactile servoing by learning from demonstration. 
Moreover, our method represents the tactile sensing information as to lie on a surface --- or a 2D manifold --- and perform a manifold learning, 
making it applicable to any tactile skin geometry. We evaluate our method on a contact point tracking task using a robot equipped with a tactile finger.
\end{abstract}


\section{Introduction}
\label{sec:introduction}
The ability to adapt actions based on tactile sensing is the key to robustly interact and manipulate with objects in the environment. Previous experiments have shown that when the tactile-driven control is impaired, humans have difficulties performing even basic manipulation tasks \cite{Johansson2018match, Johansson1984}. Hence, we believe that equipping robots with tactile feedback capability is important to make progress in robotic manipulation.


In line with this direction, recently a variety of tactile sensors \cite{wettels08, Chorley_TacTip_2009, GelSight_2017, Cheng_Tactile_Sensing_Module_2011} have been developed and used in robotics research community, and researchers have designed several tactile-driven control --- or popularly termed as \textit{tactile servoing} --- algorithms. However, many tactile servoing algorithms were designed for specific kinds of tactile sensor geometry, such as a planar surface \cite{Li2013ACF} or a spherical surface \cite{Lepora2017ETSAT}, therefore they do not apply to the broad class of tactile sensors in general. For example, if we would like to equip a robot with a tactile skin of arbitrary geometry or if there is a change in the sensor geometry due to wear or damage, we will need a more general tactile servoing algorithm.

In this paper, we present our work on a learning-based tactile servoing algorithm that does not assume a specific sensor geometry. Our method comprises three steps. At the core of our approach, we treat the tactile skin as a manifold, hence first we perform an offline neural-network based manifold learning, to learn a latent space representation which encodes the essence of the tactile sensing information. Second, we learn a latent space dynamics model from human demonstrations. Finally, we deploy our model to perform an online control --- based on both the current and target tactile signals --- for tactile servoing on a robot.


\begin{figure}[t]
    \centering
    \null\hfill
        \subfloat[Simulated Robot]{\includegraphics[width=0.163\textwidth]{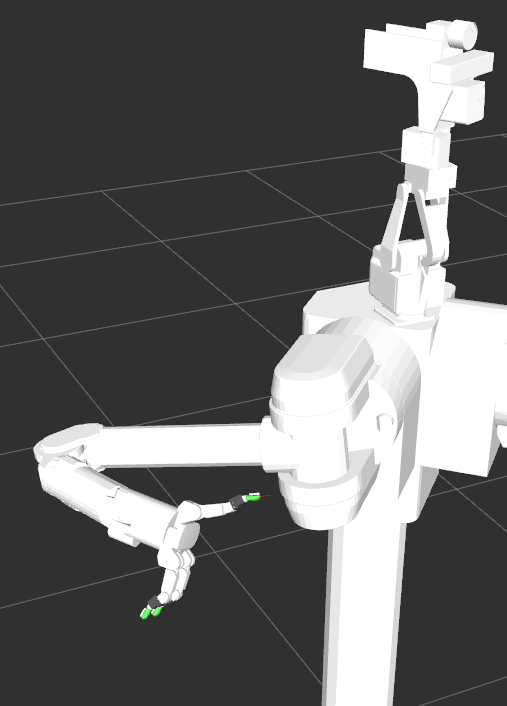}}
    \hfill
        \subfloat[Real Robot]{\includegraphics[width=0.150\textwidth,trim=0 0 0  0pt,clip]{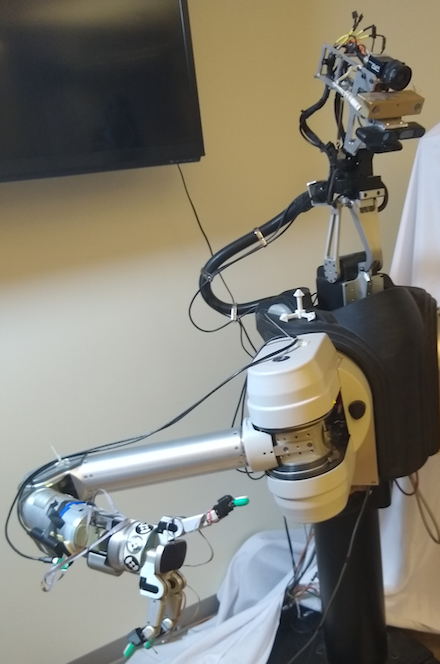}}
    \hfill
    \vspace{-0.15cm}
    \caption{\small{Learning tactile servoing platform.}}
	\label{fig:TactileServoHWsetup}
    \vspace{-0.7cm}
\end{figure}

Our contribution is twofold: First, we utilize manifold learning to impose an Euclidean structure in the latent space representation of tactile sensing, such that the control in this latent space becomes straightforward. Second, we train a single model that is able to do both forward dynamics and inverse dynamics prediction using the same demonstration dataset, which is more data-efficient than training separate models for the forward and inverse dynamics.

This paper is organized as follows. Section~\ref{sec:related_work} provides some related work. Section~\ref{sec:learning_tactile_servoing_model} presents the model that we use for learning tactile servoing from demonstration. We then present our experimental setup and evaluations in Section~\ref{sec:experiments}. Finally, we discuss our results and future work in Section~\ref{sec:discussion}.
	
\section{Related Work}
\label{sec:related_work}
	Our work is mostly inspired by previous works on learning control and dynamics in the latent space \cite{ArunSE3PoseNets, WatterEmbedToControl}. Both of these works learn a latent space representation of the state, and also learn a dynamics model in the latent space. Watter et al. \cite{WatterEmbedToControl} designed the latent space's state transition model to be locally linear, such that a stochastic optimal control algorithm can be directly applied to the learned model for control afterwards. Byravan et al. \cite{ArunSE3PoseNets} designed the latent space to represent SE(3) poses of the tracked objects in the scene, and the transition model is simply the SE(3) transformations of these poses. Control in \cite{ArunSE3PoseNets} is done by gradient-following of the Euclidean distance between the target and current latent space poses with respect to action.
\YC{Maybe say more specific what is the same and what is different in your approach.}

In this work, we train a latent space dynamics model that takes latent space representation of the current tactile sensing and applied action, and predicts the latent space representation of the next tactile sensing, which is termed as \textit{forward dynamics}. Since we use the model for control, \textit{i.e.} tactile servoing, it is also essential that we can compute actions, given both the current and target tactile sensing --- termed as \textit{inverse dynamics}. Previous work \cite{AgrawalLearningForwardAndInverseDynamics} learns separate models, one for the forward dynamics, and one for the inverse dynamics model, for a poking task. In contrast to this, in our work we train a single model for both forward and inverse dynamics.

In terms of latent space representation, our work is inspired by the work of Hadsell \textit{et al.}
\cite{HadsellDimensionalityReduction}, where they use a Siamese neural network and construct a loss function such that similar data points are close to each other in the latent space and dissimilar data points are further away from each other in the latent space. In this work, we also use a Siamese neural network, however we employ a loss function that performs Multi-Dimensional Scaling (MDS) \cite{PaiNNMDS}, such that the first two dimensions of the latent space represent the 2D map of the contact point on the tactile skin surface. Our third dimension in the latent space represents the degree of contact applied on the skin surface, \textit{i.e.} how much pressure was applied at the point of contact.

Regarding tactile servoing, besides the previous works \cite{Li2013ACF, Lepora2017ETSAT} which have been mentioned in Section~\ref{sec:introduction}, Su \textit{et al.} \cite{SuTactileServoing} designed a heuristic for tactile servoing with a tactile finger \cite{wettels08}. Our work treats the tactile sensor as a general manifold, hence the method shall apply to any tactile sensors.

Previously, learning tactile feedback has been done through reinforcement learning \cite{VanHoofTactile} or a combination of imitation learning and reinforcement learning \cite{Sung_ICRA_2017, KumarLearningDexterousManip}. Sutanto et al. \cite{icra2018_learn_tactile_feedback} learns a tactile feedback model for a trajectory-centric reactive policy. In this work, we learn a tactile servoing policy indirectly by learning a latent space dynamics model from demonstrations. As we engineer the latent space to be Euclidean by performing MDS and maintaining contact degree information, the inverse dynamics control action can be computed analytically, given both the current and target latent states. Hence, our method does not require  reinforcement learning to learn the desired behavior.

\section{Data-Driven Tactile Servoing Model}
\label{sec:learning_tactile_servoing_model}
	\subsection{Tactile Servoing Problem Formulation}
\begin{figure*}[t]
    \centering
    \null\hfill
        \subfloat[$L_{MDS}$]{\includegraphics[width=0.3\textwidth]{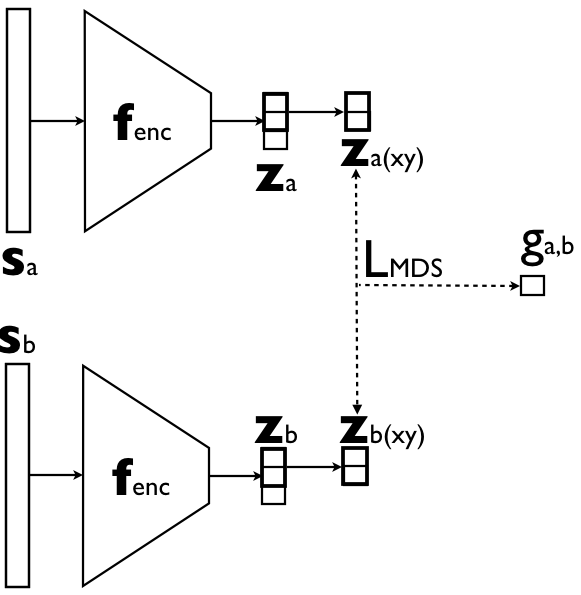}}
    \hfill
        \subfloat[$L_{AER}$, $L_{CDP}$, and $L_{LFD}$]{\includegraphics[width=0.61\textwidth]{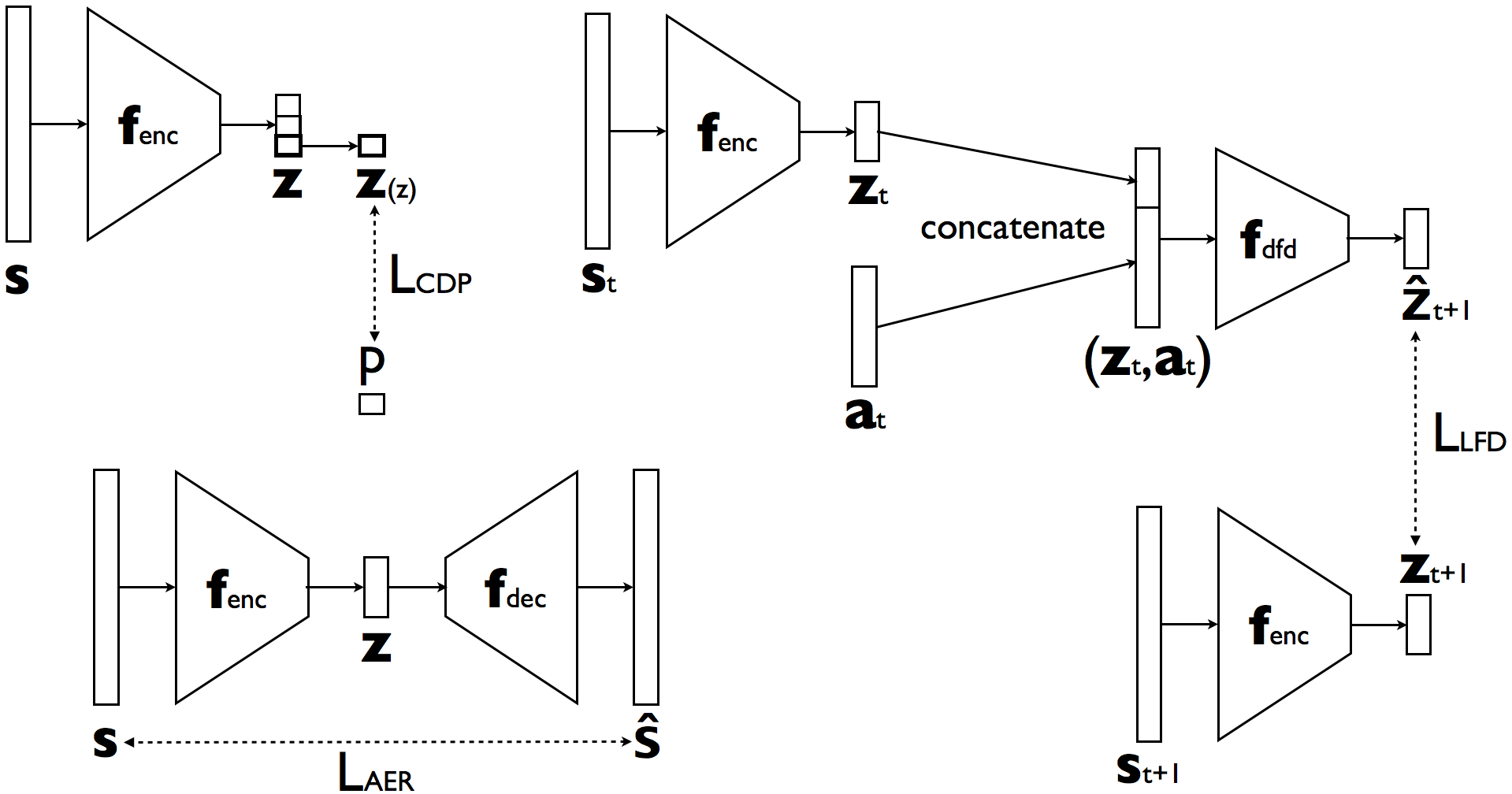}}
    \vspace{-0.15cm}
    \caption{Neural Network diagrams and its loss functions (drawn as dotted lines). Inverse dynamics loss function $L_{ID}$ is not illustrated here.}
    \label{fig:NN_diagrams}
    \vspace{-0.6cm}
\end{figure*}
Given the current tactile sensing $\sensing_t$ and the target tactile sensing $\sensing_T$, the objective is to find the action $\action_{t}$ which will bring the next tactile sensing $\sensing_{t+1} = \mathbf{f}(\sensing_t, \action_{t})$ closer to $\sensing_T$, which in the optimal case can be written as:
\begin{equation}
    \action_{t}^* = \argmin_{\action_{t}} d(\mathbf{f}(\sensing_{t}, \action_{t}), \sensing_T) 
    \label{eq:tactile_servoing_formulation}
\end{equation}

\subsection{Latent Space Representation}
\label{ssec:latent_space_rep}
If the distance metric $d$ is a squared $\mathcal{L}_2$ distance of two states, which lie on a Euclidean space and if $\mathbf{f}$ is smooth, then inverse dynamics $\action_{t}$ can be computed as proportional to $-\frac{\partial d}{\partial \action_t}$. Moreover, for some $\mathbf{f}$, the $\action_{t}^*$ in Eq. \ref{eq:tactile_servoing_formulation} can be computed analytically, at the condition $\frac{\partial d}{\partial \action_t} = \mathbf{0}$. Unfortunately, both $\sensing_{t+1}$ and $\sensing_T$ may not lie on a Euclidean space.

On the other hand, there seems to be some natural characterization of tactile sensing, such as the contact point and the degree of contact pressure applied at the point. The contact point in particular is a 3D coordinate which lies on the skin surface. 
Obviously we know that the skin surface is not Euclidean, \textit{i.e.} we cannot go from the current contact point to the target contact point by simply following the vector between them because then it may be off the skin surface while doing so\footnote{The correct way of traversing from a contact point to the other is by following the geodesics between the two points on the skin surface.}. However, if we are able to flatten the skin surface in 3D space into a 2D surface, then traversing between the two contact points translates into following the vector from one 2D point to the other on the 2D surface, which ensures that the intermediate points being traversed are all still on the 2D surface. Fortunately, there has been a method of mapping/embedding from a 3D surface into a 2D surface, called Multi-Dimensional Scaling (MDS) \cite{PaiNNMDS}. 

In this paper, we choose the latent space embedding to be three-dimensional\footnote{Here we assume that there exists a mapping from a tactile sensing $\sensing$ into the 3D contact point on the tactile skin surface as well as a mapping from $\sensing$ into the degree of contact pressure information.}:
\begin{enumerate}
    \item The first two dimensions of the latent space --called the $x$ and $y$ dimensions of the latent space-- corresponds to the 2D embedding of the 3D contact point on the tactile skin surface.
    \item The third dimension (the $z$ dimension) of the latent space represents the degree of contact pressure applied at the contact point.
\end{enumerate}
We understand that the above representation can only represent a contact as a single 3D coordinate in the latent space. Therefore, it will not be able to capture the richer set of features, such as an object's edges and orientations, etc. Tactile servoing for edge tracking is left for a future work.

\subsection{Embedding Function}
We call the latent state representation of a tactile sensing $\sensing$ as $\embedsensing$, and we define the distance metric $d$ as a squared $\mathcal{L}_2$ distance in the latent space between the embeddings of $\sensing_{t+1}$ and $\sensing_T$ by the embedding function $\embedsensing = \embedfunc(\sensing)$, as follows\footnote{Subscripts in Eq. \ref{eq:distance_metric} corresponds to time indices.}:
\begin{equation}
    d(\sensing_{t+1}, \sensing_T) = \norm{\embedsensing_{t+1} - \embedsensing_{T}}^2
    \label{eq:distance_metric}
\end{equation}
We represent the embedding function $\embedfunc$ by the encoder part of an auto-encoder neural network.

For achieving the latent space representation as mentioned in \ref{ssec:latent_space_rep}, we impose the following structure:
\begin{enumerate}
    \item We would like to map points on a surface in 3D space into 2D coordinates. Essentially this can be described as a 2D manifold embedded in 3D space. For such a manifold, the notion of distance between any pair of two 3D points on the manifold is given by the geodesics, i.e. the curve of the shortest path on the surface. For this mapping, we would like to preserve these pairwise geodesic distances in the resulting 2D map. That is, for pairs of data points $\{\{\sensing_{a}^{(1)}, \sensing_{b}^{(1)}\}, \{\sensing_{a}^{(2)}, \sensing_{b}^{(2)}\}, \dots, \{\sensing_{a}^{(K)}, \sensing_{b}^{(K)}\}\}$, we want to acquire a latent space embedding via the embedding function $\embedsensing = \embedfunc(\sensing)$ to get the latent space pairs $\{\{\embedsensing_{a}^{(1)}, \embedsensing_{b}^{(1)}\}, \{\embedsensing_{a}^{(2)}, \embedsensing_{b}^{(2)}\}, \dots, \{\embedsensing_{a}^{(K)}, \embedsensing_{b}^{(K)}\}\}$ whose distance in the $x$ and $y$ dimensions is as close as possible to the pairwise geodesic distance. Therefore we define the loss function \cite{PaiNNMDS}:
    \begin{equation}
        L_{MDS} = \sum_{k=1}^K \norm{\norm{\embedsensing_{a(xy)}^{(k)} - \embedsensing_{b(xy)}^{(k)}} - g_{a,b}^{(k)}}^2
        \label{eq:nn_mds_cost_function}
    \end{equation}
    $K$ is the number of data point pairs which is quadratic in the total number of data points $N$. $g_{a,b}^{(k)}$ is the geodesic distance between the two data points in the $k$-th pair. The pairwise geodesic distance between any two data points is approximated using the shortest path algorithm on a sparse distance matrix of $M$-nearest-neighbors of each data point. We use $M$-nearest-neighbors because the space is not 2D-Euclidean \textit{globally} due to skin curvature, but it is \textit{locally} 2D-Euclidean --i.e. flat-- on a small neighborhood (a small patch) on the skin. The computation result is stored as a symmetric dense approximate geodesic distance matrix of size $N \times N$ before the training begins. The pairwise loss function in Eq. \ref{eq:nn_mds_cost_function} is applied by using a Siamese neural network as depicted in Figure~\ref{fig:NN_diagrams}(a).
    \item Encoding of the $z$ dimension of the latent space with the contact pressure information $p$. This is done by imposing the following loss function:
    \begin{equation}
        L_{CDP} = \sum_{n=1}^N \norm{p^{(n)} - \embedsensing_{(z)}^{(n)}}^2
        \label{eq:contact_degree_info_preservation_cost_function}
    \end{equation}
\end{enumerate}
While we have the ground truth for the $z$ dimension of the latent state, i.e. $p$, we do not have the ground truth for the $x$ and $y$ dimensions. We have the 3D coordinate of the data point on the skin\footnote{For BioTacs, these 3D coordinates can be computed from electrode values, by using the point of contact estimation model presented in \cite{Lin2013EstimatingPO}.} --- which is used to compute the sparse distance matrix of M-nearest-neighbors of each data point --- but we do not know how it is mapped to the $x$ and $y$ dimensions of the latent space, and this is our reason for using an auto-encoder neural network representation. The auto-encoder reconstruction loss is:
\begin{equation}
    L_{AER} = \sum_{n=1}^N \norm{\inverseembedfunc(\embedfunc(\sensing^{(n)})) - \sensing^{(n)}}^2
    \label{eq:ae_reconstruction_loss}
\end{equation}
with $\embedfunc$ is the encoder/embedding function, and $\inverseembedfunc$ is the decoder/inverse-embedding function.

\subsection{Latent Space Forward Dynamics (LFD)}
\label{ss:latent_space_fwd_dyn}
We assume the latent space forward dynamics as follows:
\begin{equation}
    \dot{\embedsensing}_{t} = \latentdynfunc(\embedsensing_{t}, \action_{t}; \dynmodelnnparams)
    \label{eq:latent_space_dyn_model}
\end{equation}
where $\dynmodelnnparams$ is the set of trainable (neural network) parameters of the dynamics model. Numerical integration gives us the discretized version:
\begin{align*}
    \embedsensing_{t+1} = \embedsensing_{t} + \dot{\embedsensing}_{t} \deltat &= \embedsensing_{t} + \latentdynfunc(\embedsensing_{t}, \action_{t}; \dynmodelnnparams) \deltat \\
    &= \discretelatentdynfunc(\embedsensing_{t}, \action_{t}, \deltat; \dynmodelnnparams) \numberthis
    \label{eq:latent_space_integration}
\end{align*}
There are two possibilities of $\latentdynfunc$ as follows:
\subsubsection{Locally Linear (LL) LFD Model}
\label{sss:locally_linear_latent_space_fwd_dyn}
\begin{equation}
    \dot{\embedsensing}_{t} = \latentdynfunc(\embedsensing_{t}, \action_{t}; \dynmodelnnparams) = \linearlatentdynA_{t} \embedsensing_{t} + \linearlatentdynB_{t} \action_{t} + \linearlatentdync_{t}
    \label{eq:locally_linear_latent_space_dyn_model}
\end{equation}
where $\linearlatentdynA_{t}$, $\linearlatentdynB_{t}$, and $\linearlatentdync_{t}$ is predicted by a fully-connected neural network (fcnn) from input $\embedsensing_{t}$, as follows:
\begin{equation}
    \begin{bmatrix}
        \text{vec}(\linearlatentdynA_{t})\\
        \text{vec}(\linearlatentdynB_{t})\\
        \linearlatentdync_{t}
    \end{bmatrix} = \mathbf{h}_{fcnnLL}(\embedsensing_{t}; \dynmodelnnparams)
    \label{eq:locally_linear_param_prediction}
\end{equation}
with $\text{vec}(\linearlatentdynA_{t})$ and $\text{vec}(\linearlatentdynB_{t})$ are the vectorized representation of $\linearlatentdynA_{t}$ and $\linearlatentdynB_{t}$, respectively \cite{WatterEmbedToControl}.
\subsubsection{Non-Linear (NL) LFD Model}
\label{sss:non_linear_latent_space_fwd_dyn}
\begin{equation}
    \dot{\embedsensing}_{t} = \latentdynfunc(\embedsensing_{t}, \action_{t}; \dynmodelnnparams) = \mathbf{h}_{fcnnNL}(
    \begin{bmatrix}
        \embedsensing_{t}\\
        \action_{t}
    \end{bmatrix}; \dynmodelnnparams)
    \label{eq:non_linear_latent_space_dyn_model}
\end{equation}
We would like to be able to predict the forward dynamics in the latent space, so we use the following loss function:
\begin{equation}
    L_{LFD} = \sum_{t=1}^{H} \norm{\discretelatentdynfunc(\embedsensing_{t}, \action_{t}, \deltat; \dynmodelnnparams) - \embedfunc(\sensing_{t+1})}^2
    \label{eq:latent_forward_dyn_loss}
\end{equation}
with $\embedsensing_{t+1}$ is computed from Eq. \ref{eq:latent_space_dyn_model} and \ref{eq:latent_space_integration}. For additional robustness, we can also do chained predictions for $C$ time steps ahead and sum up the loss function in Eq. \ref{eq:latent_forward_dyn_loss} for these chains, similar to the work by Nagabandi et al. \cite{NagabandiChainPrediction}.

\subsection{Inverse Dynamics (ID)}
\label{ss:id_definition}
Beside forward dynamics, we also found that the ability of the model to predict inverse dynamics to be essential for the purpose of action selection or control.

There are three possibilities of inverse dynamics model:
\subsubsection{Locally Linear (LL) ID}
Based on the locally linear LFD model \cite{WatterEmbedToControl} in section \ref{sss:locally_linear_latent_space_fwd_dyn}, Eq. \ref{eq:latent_space_dyn_model}, \ref{eq:latent_space_integration}, \ref{eq:locally_linear_latent_space_dyn_model}, we can setup a constrained optimal control problem:
\begin{mini}|s|
	{\action_{t}, \embedsensing_{t+1}}{\frac{1}{2} \norm{\embedsensing_{T} - \embedsensing_{t+1}}^2 + \frac{\beta}{2} \norm{\action_{t}}^2}
	{}{}
	\addConstraint{\embedsensing_{t+1}}{= \embedsensing_{t} + (\linearlatentdynA_{t} \embedsensing_{t} + \linearlatentdynB_{t} \action_{t}} + 
	\linearlatentdync_{t}) \deltat
	\label{eq:optimal_control_formula_of_linear_latent_space_dynamics}
\end{mini}
whose solution is:
\begin{equation}
    \action_{t,ID} = \linearlatentdynB_{t}\T \left(\linearlatentdynB_{t} \linearlatentdynB_{t}\T + \frac{\beta}{{\deltat}^2} \eye\right)^{-1} \left(\frac{\embedsensing_{T} - \embedsensing_{t}}{\deltat} - \linearlatentdynA_{t} \embedsensing_{t} - \linearlatentdync_{t}\right)
    \label{eq:locallylinearID_optimal_control_solution}
\end{equation}
\subsubsection{Negative-Gradient (NG) ID}
Based on the non-linear LFD model in section \ref{sss:non_linear_latent_space_fwd_dyn}, we can compute a gradient-based controller which minimizes the distance function $d = \norm{\discretelatentdynfunc(\embedsensing_{t}, \action_{t}, \deltat; \dynmodelnnparams) - \embedsensing_{T}}^2$, that is \cite{ArunSE3PoseNets}:
\begin{equation}
    \action_{t,ID} = - \alpha \frac{\partial d}{\partial \action_{t}}\biggr\rvert_{\action_{t}=\mathbf{0}}
    \label{eq:nnneggradientID_optimal_control_solution}
\end{equation}
with $\alpha$ is a positive contant that scales the gradient w.r.t. maximum allowed magnitude of $\action_{t}$

\subsubsection{Neural Network Jacobian (NJ) ID}
Based on the non-linear LFD model in section \ref{sss:non_linear_latent_space_fwd_dyn}, we can derive the following from Eq. \ref{eq:non_linear_latent_space_dyn_model} (dropping time index $t$ for a moment):
\begin{equation}
    \ddot{\embedsensing} = 
    \begin{bmatrix}
        \jacembedsensing & \jacaction
    \end{bmatrix}
    \begin{bmatrix}
        \dot{\embedsensing}\\
        \dot{\action}
    \end{bmatrix} = \jacembedsensing \dot{\embedsensing} + \jacaction \dot{\action}
    \label{eq:tangent_space_formula_of_nonlinear_latent_space_dynamics}
\end{equation}
where $\jacembedsensing$ and $\jacaction$ are the Jacobians of $\mathbf{h}_{fcnnNL}$ w.r.t. $\embedsensing$ and $\action$, respectively,
which can be discretized into:
\begin{align*}
    \frac{(\embedsensing_{t+1} - \embedsensing_{t}) - (\embedsensing_{t} - \embedsensing_{t-1})}{\deltat^2} = 
    &\jacprevembedsensing \frac{\embedsensing_{t} - \embedsensing_{t-1}}{\deltat}\\
    &+ \jacprevaction \frac{\action_{t} - \action_{t-1}}{\deltat} \numberthis
\end{align*}
or
\begin{align*}
    \embedsensing_{t+1} = &\embedsensing_{t} + \left( \frac{1}{\deltat} \eye + \jacprevembedsensing \right) \deltat \left( \embedsensing_{t} - \embedsensing_{t-1} \right)\\ 
    &+ \jacprevaction \deltat \left( \action_{t} - \action_{t-1} \right) \numberthis
    \label{eq:linearized_nonlinear_latent_space_dynamics}
\end{align*}
with $\jacprevembedsensing$ and $\jacprevaction$ are the Jacobians of $\mathbf{h}_{fcnnNL}$\footnote{These Jacobians exist in our experiment because we choose smooth activation functions for $\mathbf{h}_{fcnnNL}$, such as hyperbolic tangent (tanh).} w.r.t. previous latent state $\embedsensing_{t-1}$ and previous action $\action_{t-1}$, respectively. Let us define $\prevlinearizednonlinlatentdynA = \left( \frac{1}{\deltat} \eye + \jacprevembedsensing \right)$, $\prevlinearizednonlinlatentdynB = \jacprevaction$, and $\prevlinearizednonlinlatentdync = -\prevlinearizednonlinlatentdynA \embedsensing_{t-1} - \prevlinearizednonlinlatentdynB \action_{t-1}$, Eq. \ref{eq:linearized_nonlinear_latent_space_dynamics} can be written as:
\begin{equation}
    \embedsensing_{t+1} = \embedsensing_{t} + (\prevlinearizednonlinlatentdynA \embedsensing_{t} + \prevlinearizednonlinlatentdynB \action_{t} + \prevlinearizednonlinlatentdync) \deltat
    \label{eq:linearized_nonlinear_latent_space_dynamics_termed}
\end{equation}
We can setup a constrained optimal control problem:
\begin{mini}|s|[3]
	{\action_{t}, \embedsensing_{t+1}}{\frac{1}{2} \norm{\embedsensing_{T} - \embedsensing_{t+1}}^2 + \frac{\beta}{2} \norm{\action_{t}}^2}
	{}{}
	\addConstraint{\embedsensing_{t+1}}{= \embedsensing_{t} + (\prevlinearizednonlinlatentdynA \embedsensing_{t} + \prevlinearizednonlinlatentdynB \action_{t} + \prevlinearizednonlinlatentdync) \deltat}
	\label{eq:optimal_control_formula_of_linearized_nonlinear_latent_space_dynamics}
\end{mini}
whose solution is:
\begin{align*}
    \action_{t,ID} = \prevlinearizednonlinlatentdynB\T \left(\prevlinearizednonlinlatentdynB \prevlinearizednonlinlatentdynB\T + \frac{\beta}{{\deltat}^2} \eye\right)^{-1} \Big(&\frac{\embedsensing_{T} - \embedsensing_{t}}{\deltat} - \prevlinearizednonlinlatentdynA \embedsensing_{t} \\
    &- \prevlinearizednonlinlatentdync\Big) \numberthis
    \label{eq:nnjacobianID_optimal_control_solution}
\end{align*}

The optimal control formulation in Eq. \ref{eq:optimal_control_formula_of_linear_latent_space_dynamics} and \ref{eq:optimal_control_formula_of_linearized_nonlinear_latent_space_dynamics} are similar to those of Linear Quadratic Regulator (LQR) with (finite) horizon equal to 1. 
Derivations of Eq. \ref{eq:locallylinearID_optimal_control_solution} and \ref{eq:nnjacobianID_optimal_control_solution} from Eq. \ref{eq:optimal_control_formula_of_linear_latent_space_dynamics} and \ref{eq:optimal_control_formula_of_linearized_nonlinear_latent_space_dynamics}, respectively, can be seen in the Appendix.

From Eq. \ref{eq:locallylinearID_optimal_control_solution}, \ref{eq:nnneggradientID_optimal_control_solution}, and \ref{eq:nnjacobianID_optimal_control_solution}, in general we can write:
\begin{equation}
    \action_{t,ID} = \invdynfunc(\embedsensing_{T}, \embedsensing_{t}, \embedsensing_{t-1}, \action_{t-1}, \deltat; \dynmodelnnparams)
    \label{eq:general_ID_solution_function}
\end{equation}
Please note that $\dynmodelnnparams$ are shared between $\discretelatentdynfunc$ (Eq. \ref{eq:latent_space_integration}) and $\invdynfunc$ (Eq. \ref{eq:general_ID_solution_function}).

For our purpose, it is mostly important that the inferred inverse dynamics action points to the right direction. Therefore, we can leverage the demonstration dataset to also optimize the following inverse dynamics loss:
\begin{align*}
    &L_{ID} = \\
    &\sum_{t=1}^{H} \norm{\frac{\invdynfunc(\embedsensing_{T}=\embedsensing_{t+1}, \embedsensing_{t}, \embedsensing_{t-1}, \action_{t-1}, \deltat; \dynmodelnnparams)}{\norm{\invdynfunc(\embedsensing_{T}=\embedsensing_{t+1}, \embedsensing_{t}, \embedsensing_{t-1}, \action_{t-1}, \deltat; \dynmodelnnparams)}} -  \frac{\action_t}{\norm{\action_t}}}^2 \numberthis
    \label{eq:inverse_dyn_loss}
\end{align*}
We combine the loss functions as follows:
\begin{align*}
    L_{totalAE} &= w_{AER} L_{AER} + w_{MDS} L_{MDS} + w_{CDP} L_{CDP} \numberthis \\
    L_{totalDyn} &= w_{LFD} L_{LFD} + w_{ID} L_{ID} \numberthis
    \label{eq:total_loss_function}
\end{align*}
with the weights $w_{MDS}, w_{CDP}, w_{AER}, w_{LFD}, w_{ID}$ are tuned so that each loss function components become comparable to each other. Some individual loss functions are depicted in Figure~\ref{fig:NN_diagrams}.
$L_{totalAE}$ and $L_{totalDyn}$ are minimized separately (with separate optimizer) in parallel with respect to the human demonstrations' trajectory dataset $\{(\sensing_{t-1}, \action_{t-1}, \sensing_{t}, \action_{t}, \sensing_{t+1})\}_{t \in \{1,\dots,H\}}$. Minimizing $L_{totalDyn}$ effectively means to train the dynamics model parameters $\dynmodelnnparams$ to minimize both the forward dynamics loss $L_{LFD}$ and the inverse dynamics loss $L_{ID}$.

\section{Experiments}
\label{sec:experiments}
	\subsection{Experimental Setup}
We use the right arm of a bi-manual anthropomorphic robot system, which is a 7-degrees-of-freedom Barrett WAM arm, plus its hand which has three fingers. We mount a biomimetic tactile sensor BioTac \cite{wettels08} on the tip of the middle finger of the right hand. The finger joints configuration are programmed to be fixed during demonstration and testing. The setup is pictured in Figure \ref{fig:TactileServoHWsetup}. We setup the end-effector frame to coincide with the BioTac finger frame as described in \cite{Lin2013EstimatingPO}, figure 4. The BioTac has 19 electrodes distributed on the skin surface, capable of measuring deformation of the skin by measuring the change of impedance when the conductive fluid underneath the skin is being compressed or deformed due to a contact with an object. In our experiments, $\sensing$ is a vector of 19 values corresponding to the digital reading of the 19 electrodes, subtracted with its offset value estimated when the finger is in the air and does not make any contact with any object. The contact pressure information $p$ is a scalar quantity, which is obtained by negating the mean of the vector $\sensing$, i.e. $p = -\bar{\sensing} = -\frac{1}{19} \sum_{i=1}^{19} s_i$, with $s_i$ being the digital reading of the $i$-th electrode minus its offset.

\subsubsection{Human Demonstration Collection}
For collecting human demonstrations, we set the robot to be in a gravity compensation mode, allowing a human demonstrator to guide the robot to a sequence of contact interaction between the BioTac finger and a drawer handle. The robot's sampling and control frequency is 300 Hz, while the tactile information $\sensing$ is sampled at 100 Hz. Later $p$ can be computed from $\sensing$.

The demonstrations are split into two parts: one part corresponds to the contact interaction dynamics due to the rotational change of the finger pose, and the other part due to the translational change of the finger pose. Each part comprises of 7 sub-parts which correspond to contact point trajectories that traverse through different areas of the skin. For each sub-part of the rotational motion, we provide 3 demonstrations, while for the translational motion, we provide 4 demonstrations. These are determined such that we have a 50\%-50\% composition of data points for rotational and translational motion, respectively. Each rotational demonstration involves the sequence of making contact, rotational motion clock-wise w.r.t. x-axis of the finger frame, breaking contact, making contact again, rotational motion counter-clock-wise, and finally breaking contact. Each translational demonstration involves the sequence of making contact, swiping motion forward on the x-axis of the finger frame, breaking contact, making contact again, swiping motion backward, breaking contact again, and then repeat the whole sequence one more time. The breaking and making contacts are intentionally done to make data segmentation easier, by using a zero-crossing algorithm~\cite{fod2002automated}.

In total we collect $N=55431$ data points of the tactile sensing vector $\sensing$, and $H=183825$ pairs of $(\sensing_{t-1}, \action_{t-1}, \sensing_{t}, \action_{t}, \sensing_{t+1})$. Instead of constructing a single massive geodesic distance matrix of size $N \times N$, we split the data randomly into $P$ bins, each of size $N'=2310$, so we end up with $P$ geodesic distance matrices, each of size $N' \times N'$.  During training, for each siamese pair being picked, both data points must be associated with the same geodesic distance matrix. For the geodesic distance computation, we use a nearest-neighborhood of size $M=18$. On the other hand, we obtain the number of state-action pairs $H$ after excluding the pairs that contain states which correspond to contact pressure information $p$ below a specific threshold. We exclude these pairs as we deem them being off-contact tactile states and not being informative for performing tactile servoing\footnote{In the extreme case, when the robot is not in contact with any object, there is no point of performing tactile servoing.}.

After collecting the demonstrations, we pre-process the data by performing low-pass filtering with a cut-off frequency of 1 Hz. We determined this cut-off frequency by visualizing the frequency-domain analysis of the data. This frequency selection of tactile servoing is also supported by a previous work by Johansson et al. \cite{JohanssonCoding}. During training, we perform the forward dynamics prediction at $\frac{100}{29}$, $\frac{100}{30}$, $\frac{100}{31}$, $\frac{100}{32}$, and $\frac{100}{33}$ Hz, while during testing, the model predict at $100/31$ Hz. In general it is hard to predict at higher frequencies, because demonstrations are performed slowly.

\GS{Maybe this number of pairs need to be reorganized and moved to after action representation; we also need to explain that we perform low pass filtering at 5 Hz and state forward dynamics prediction at 10 Hz by re-organizing the data and averaging in-between actions}

\subsubsection{Action Representation}
We choose the end-effector velocity expressed with respect to the end-effector frame as the action/policy representation. By representing the end-effector velocity with respect to the end-effector frame, effectively we are cancelling out the dependency of the state representation on the end-effector pose information, making the learned policy easier to generalize to new situations. Moreover, this choice also naturally takes care of repeatable position tracking error of the end-effector.

We use the Simulation Lab (SL) robot control framework \cite{Schaal_SL_2009} in our experiments\footnote{In the previous version of our experiment we use Riemannian Motion Policies (RMP) \cite{RatliffRMP} for the robot control framework.}. The framework provides us with the end-effector velocity with respect to the robot base frame $\cartesianvelocity_b$. To get the end-effector velocity with respect to the end-effector frame $\cartesianvelocity_e$, we compute the following~\cite{SicilianoRoboticsTextbook}:
\begin{equation}
    \cartesianvelocity_e = 
    \begin{bmatrix}
        \mathbf{R}_e\T & \mathbf{0} \\
        \mathbf{0} & \mathbf{R}_e\T
    \end{bmatrix}
    \cartesianvelocity_b
    \label{eq:base_to_endeff_velocity_conversion}
\end{equation}
where $\mathbf{R}_e$ is the end-effector orientation with respect to the base frame, expressed as a rotation matrix.
Hence, we define the action $\action = \cartesianvelocity_e$ with dimensionality 6, where the first three dimensions is the linear velocity and the last three is the angular velocity. During the demonstration, the robot is sampled at 300 Hz, but the prediction is made at 3 Hz, for this we summarize by averaging all $\cartesianvelocity_b$'s applied between $\sensing_t$ and $\sensing_{t+1}$, and then convert this average to $\cartesianvelocity_e$.

\subsubsection{Machine Learning Framework and Training Process}
Our auto-encoder takes in 19 dimensional input vector $\sensing$, and compresses it down to a 3 dimensional latent state embedding, $\embedsensing$. The intermediate hidden layers are fully connected layers of size 19, 12, 6, all with \textit{tanh} activation function, forming the encoder function $\embedfunc$. The decoder part is a mirrored structure of the encoder function, forming $\inverseembedfunc$. $\mathbf{h}_{fcnnNL}$ is a feedforward neural network with 9 dimensional input (3 dimensional latent state $\embedsensing$ and 6 dimensional action policy $\action$), 1 hidden layer of size 15 with tanh activation functions, and 3 dimensional output. $\mathbf{h}_{fcnnLL}$ is a feedforward neural network with 3 dimensional latent state $\embedsensing$ as input, 3 hidden layers of size 8, 15, 23, all with tanh activation function, and 30 dimensional output which corresponds to the parameters of $\linearlatentdynA_{t}$, $\linearlatentdynB_{t}$, and $\linearlatentdync_{t}$ in Eq. \ref{eq:locally_linear_param_prediction}.

We use a batch size of 128, and we use separate RMSProp optimizers \cite{Tieleman2012} to minimize $L_{totalAE}$ and $L_{totalDyn}$ for 200k iterations. We set the values of $w_{MDS} = 2 \times 10^7$, $w_{CDP}=2 \times 10^7$, $w_{AER}=100$, $w_{LFD}=1 \times 10^8$, $w_{ID}=1 \times 10^7$, and $\beta=0.1$ empirically. We implement all components of our model in TensorFlow \cite{TensorFlowBib}. We also noticed a significant improvement in learning speed and fitting quality after we add Batch Normalization \cite{Ioffe_BatchNorm} layers in our model.

\subsection{Auto-Encoder Reconstruction Performance}
Our first evaluation is on the reconstruction performance of the auto-encoder in terms of normalized mean squared error (NMSE). NMSE is the mean squared prediction error divided by the variance of the ground truth. We obtain all NMSE values are below 0.25 for the training (85\% split), validation (7.5\% split), and test (7.5\% split) sets.

\subsection{Neural Network  Multi-Dimensional Scaling (MDS)}
\begin{figure}[ht]
    \vspace{-0.2cm}
	\centering
    \includegraphics[width=\columnwidth,trim={0 10pt 0 50pt},clip]{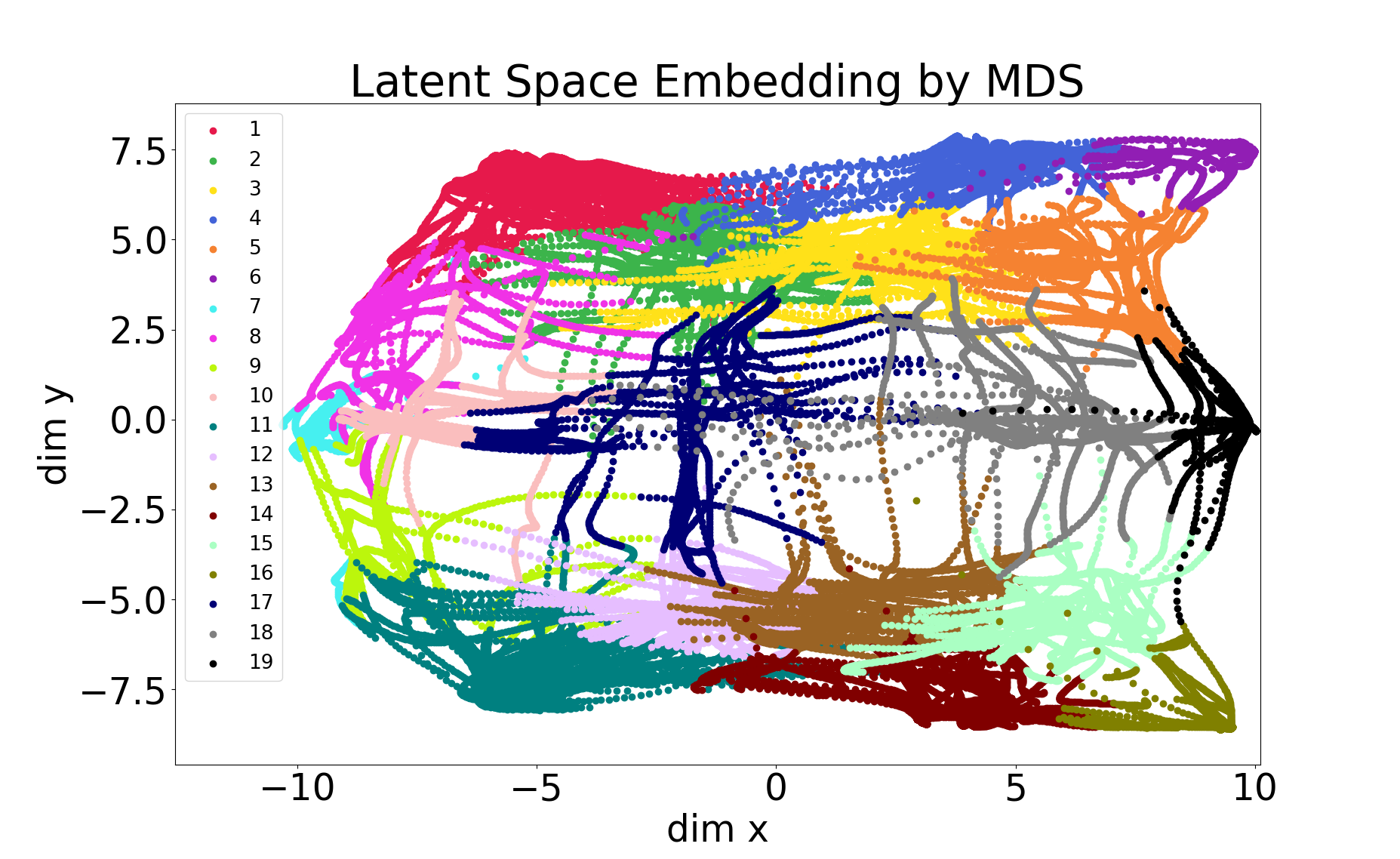}
    \vspace{-0.5cm}
	\caption{\small{x-y dimensions of latent space embedding by MDS}}
	\label{fig:nnmds_xy_embedding}
    \vspace{-0.3cm}
\end{figure}

In terms of MDS performance, we plot the x-y coordinates of the latent space embedding of all tactile sensing $\sensing$ data points in the demonstration data, in Figure \ref{fig:nnmds_xy_embedding}. Each data point is colored and labeled based on the BioTac electrode index with maximum activation. This result agrees with the Figure 2 of \cite{Lin2013EstimatingPO}.
Moreover, we randomly sampled 10000 siamese pairs from the training, validation, and test dataset, and compare their x-y Euclidean distance in the latent space vs. the ground truth geodesic distance. We got all these comparisons to have NMSE less than 0.02.

\subsection{Latent Forward Dynamics Prediction Performance}
\begin{figure}[ht]
    \vspace{-0.4cm}
	\centering
    \includegraphics[width=\columnwidth,trim={0 0pt 0 20pt},clip]{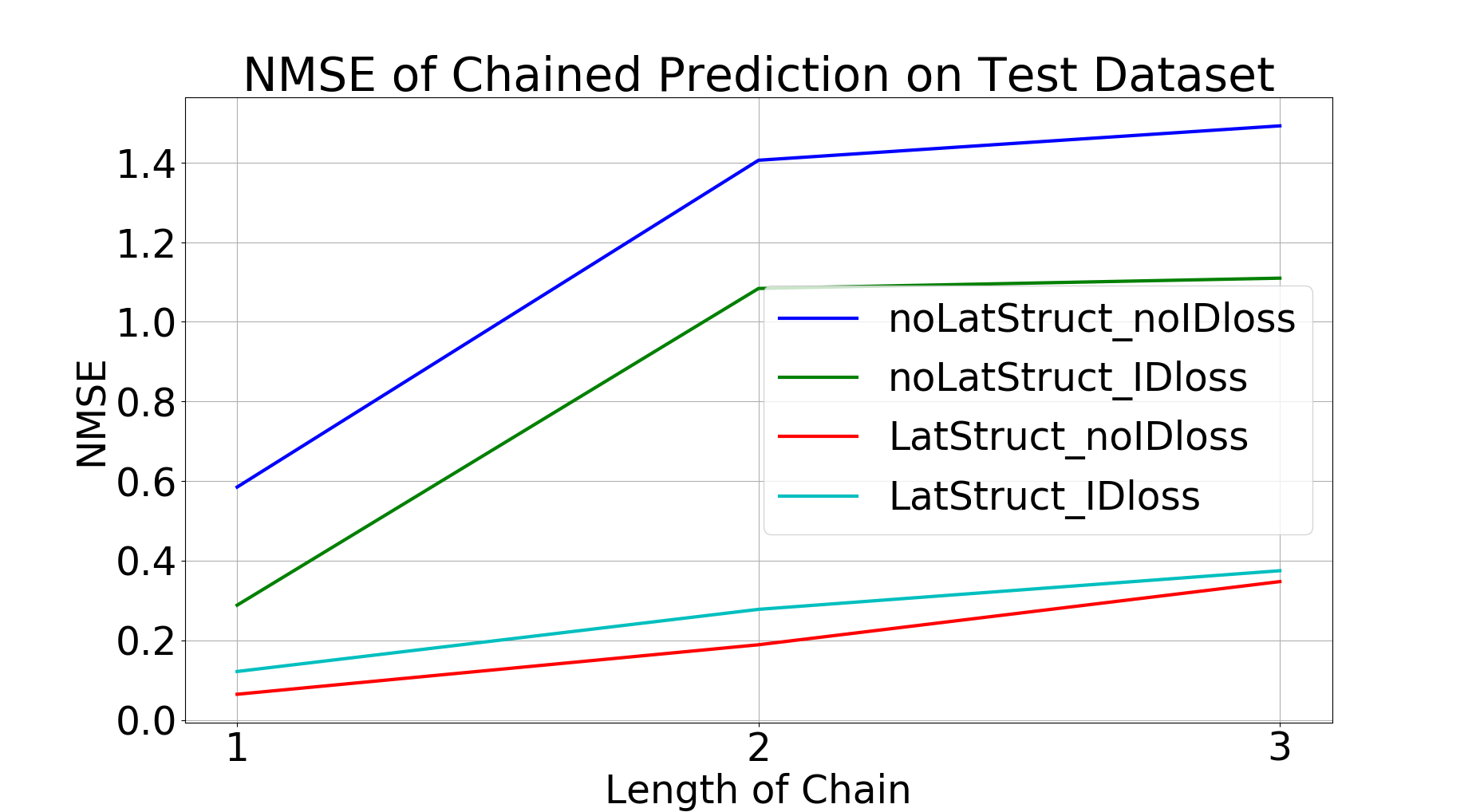}
    \vspace{-0.6cm}
	\caption{\small{Normalized mean squared error (NMSE) vs. the length of chained forward dynamics prediction, averaged over latent space dimensions, on test dataset.}}
	\label{fig:chained_prediction_nmse}
    \vspace{-0.6cm}
\end{figure}

We trained the latent space forward dynamics function $\latentdynfunc$ by chain-predicting the next $C$ latent states with a length of training chain $C_{train}=2$ and testing it with a length of chain $C_{test}=3$. We then evaluate the NMSEs. In Figure~\ref{fig:chained_prediction_nmse}, we compare the performance between 4 different combinations:
\begin{itemize}
    \item using both of $L_{MDS}$ and $L_{CDP}$ loss functions during training as indicated by \textit{LatStruct} or not using both of them --i.e. without any structure imposed in the latent space representation-- as indicated by \textit{noLatStruct}, and
    \item using inverse dynamics loss $L_{ID}$ during training as indicated by \textit{IDloss} or not using it (\textit{noIDloss}).
\end{itemize}
We see that in all cases where no latent space structure is imposed, the performance is generally worse than those with imposed latent space structure. We believe this happens because it is a hard task to train a forward dynamics predictor to predict on an unstructured latent space. On the other hand, in general we see that all models with imposed inverse dynamics loss $L_{ID}$ perform worse than those without $L_{ID}$. We think this is most likely because training a model without imposing $L_{ID}$ loss is easier than training with imposing it. However, as we will see in section~\ref{ss:id_evaluation}, the model trained without $L_{ID}$ loss does not provide correct action policies for tactile servoing as it was not trained to do so.

\subsection{Inverse Dynamics Prediction Performance}
\label{ss:id_evaluation}
\begin{figure}[ht]
    \vspace{-0.5cm}
	\centering
    \includegraphics[width=\columnwidth,trim={0 50pt 0 40pt},clip]{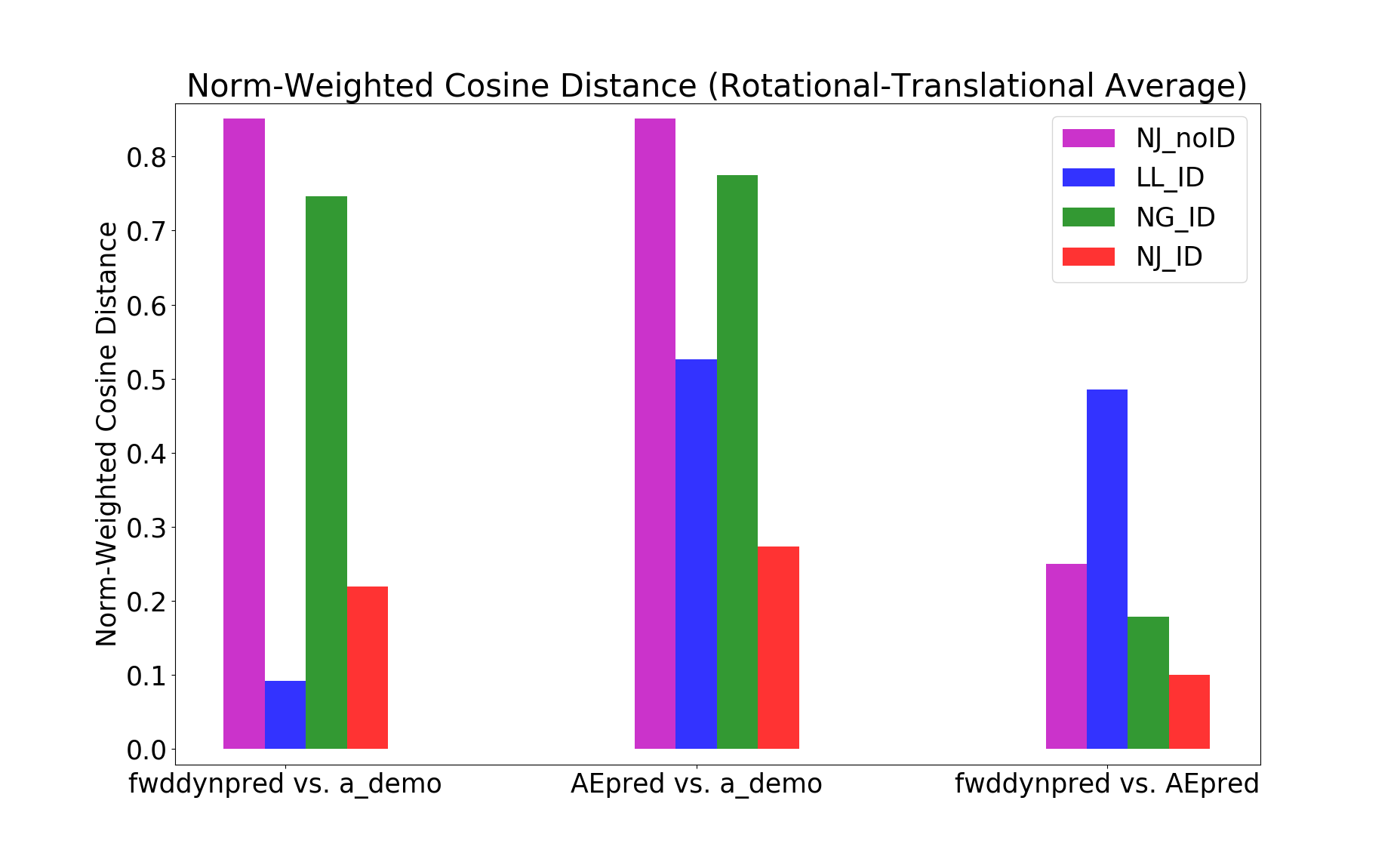}
    \vspace{-0.5cm}
	\caption{\small{Average cosine distance between rotational and translational inverse dynamics, weighted by the norm of the ground truth.}}
	\label{fig:invdyn_cosine_distance_comparison}
    \vspace{-0.3cm}
\end{figure}
In Figure \ref{fig:invdyn_cosine_distance_comparison}, we compare the inverse dynamics prediction performance between the 3 possible inverse dynamics models as described in section \ref{ss:id_definition}, in terms of the average cosine distance between rotational and translational\footnote{Translational and rotational components here correspond to the first three and the last three dimensions of $\action$, respectively.} inverse dynamics, weighted by the norm of the ground truth. We termed \textit{fwddynpred} and \textit{AEpred} for inverse dynamics prediction in Eq. \ref{eq:general_ID_solution_function} by setting $\embedsensing_T$ equal to the \textit{constant} value of $\discretelatentdynfunc(\embedsensing_{t}, \action_{t}, \deltat; \dynmodelnnparams)$ and $\embedfunc(\sensing_{t+1})$, respectively. Obviously $\discretelatentdynfunc(\embedsensing_{t}, \action_{t}, \deltat; \dynmodelnnparams)$ is an easier target for inverse dynamics than $\embedfunc(\sensing_{t+1})$, as apparent from the better prediction performance of the left bar group as compared to the middle bar group. If $\latentdynfunc$ is trained well, we can expect that the performance between \textit{fwddynpred} and \textit{AEpred} becomes more similar. On the right bar group, we also compare \textit{fwddynpred} vs. \textit{AEpred}: poor performance here indicate whether $\latentdynfunc$ could not predict well, or unstable $\invdynfunc$ (i.e. a big change in $\action_{t,ID}$ for a small change in $\embedsensing_T$). With respect to this analysis, we deem Neural Network Jacobian (NJ) to be the best inverse dynamics model.
We also evaluate \textit{NJ\_noID} which corresponds to not minimizing $L_{ID}$ loss. By comparing \textit{NJ\_noID} and \textit{NJ\_ID}, we can see that minimizing $L_{ID}$ loss is essential for acquiring a good inverse dynamics model.

\subsection{Real Robot Experiment}
\label{ss:real_robot_exp_tactile_servoing}
\begin{figure}[ht]
    \vspace{-0.8cm}
    \centering
    \null\hfill
        \subfloat[Rotation 1]{\includegraphics[width=0.10125\textwidth,trim={0 175pt 0 70pt},clip]{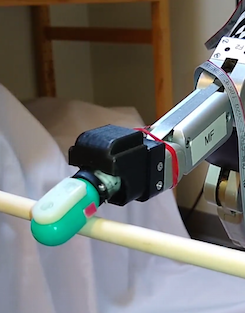}}
    \hfill
        \subfloat[Rotation 2]{\includegraphics[width=0.10125\textwidth,trim={0 10pt 0  135pt},clip]{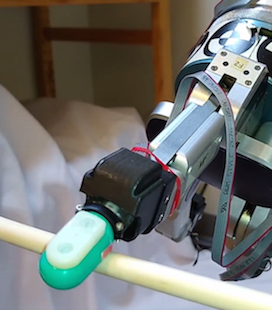}}
    \hfill
        \subfloat[Rotation 3]{\includegraphics[width=0.10125\textwidth,trim={2pt 0 2pt 150pt},clip]{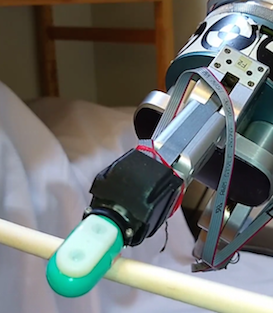}}
    \hfill
        \subfloat[Rotation 4]{\includegraphics[width=0.10125\textwidth,trim={0 7pt 0  190pt},clip]{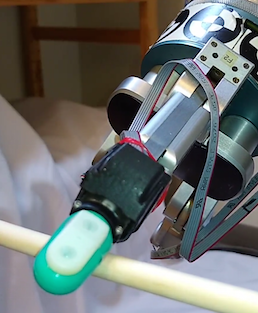}}
    \hfill\null
    \par
    \vspace{-0.2cm}
    \null\hfill
        \subfloat[Translat. 1]{\includegraphics[width=0.10125\textwidth,trim={0 0pt 0 250pt},clip]{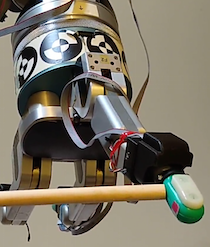}}
    \hfill
        \subfloat[Translat. 2]{\includegraphics[width=0.10125\textwidth,trim={0 0pt 0 230pt},clip]{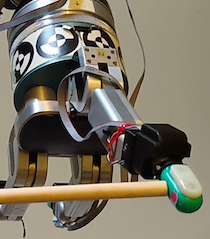}}
    \hfill
        \subfloat[Translat. 3]{\includegraphics[width=0.10125\textwidth,trim={0pt 0pt 0 240pt},clip]{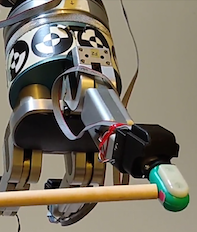}}
    \hfill
        \subfloat[Translat. 4]{\includegraphics[width=0.10125\textwidth,trim={0 0pt 0 205pt},clip]{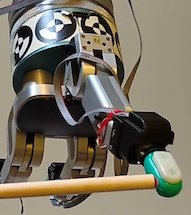}}
    \hfill\null
    \caption{Snapshots of our experiments executing the tactile servoing with the learned model (non-linear LFD model and neural network Jacobian ID model) on a real robot. Red sticker indicates the target contact point. The first row, figures (a)-(d) are for a target contact point whose achievement requires rotational change of pose of the BioTac finger. The second row, figures (e)-(h) are for a target contact point whose achievement requires translational change of pose of the BioTac finger.}
    \label{fig:tactile_servoing_executions}
    \vspace{-0.3cm}
\end{figure}
%
In Fig. ~\ref{fig:tactile_servoing_executions}, we provide snapshots of robot executions on a real hardware\footnote{The model gives $\cartesianvelocity_e$ as output, while the robot only knows how to track $\cartesianvelocity_b$, thus we need to invert Eq.~\ref{eq:base_to_endeff_velocity_conversion} to perform tactile servoing.} with real-time tactile sensing from the BioTac finger. We see that the system is able to produce the required rotational motions (Fig. ~\ref{fig:tactile_servoing_executions} (a)-(d)) and translational motions (Fig. ~\ref{fig:tactile_servoing_executions} (e)-(h)) needed to achieve the specified target contact point.
The full pipeline of the experiment can be seen in the video \url{https://youtu.be/0QK0-Vx7WkI}.

\section{Discussion and Future Work}
\label{sec:discussion}
	In this work, we presented a learning-from-demonstration framework for achieving tactile servoing behavior. We showed that our manifold representation learning of tactile sensing information is critical to the success of our approach. We also showed that for learning a tactile servoing model, it is important to not only be able to predict the next state from the current state and action (forward dynamics), but also be able to predict the action if given a target state (inverse dynamics).

In the future, we would like to extend our work to not only track a contact point, but also a contact profile surrounding the point. This can be useful to produce interesting behaviors such as tactile navigation on the edges of an object.




\section*{APPENDIX}
\label{sec:appendix}
    Given the constrained optimization problem:
\begin{mini}|s|
	{\action_{t}, \embedsensing_{t+1}}{\frac{1}{2} \norm{\embedsensing_{T} - \embedsensing_{t+1}}^2 + \frac{\beta}{2} \norm{\action_{t}}^2}
	{}{}
	\addConstraint{\embedsensing_{t+1}}{= \embedsensing_{t} + (\linearlatentdynA_{t} \embedsensing_{t} + \linearlatentdynB_{t} \action_{t} + \linearlatentdync_{t}) \deltat}
	\label{eq:optimal_control_problem}
\end{mini}
which is equivalent to solving the following problem:
\begin{mini}
	{\action_{t}, \embedsensing_{t+1}, \lagrangemultiplier}{\frac{1}{2} \norm{\embedsensing_{T} - \embedsensing_{t+1}}^2 + \frac{\beta}{2} \norm{\action_{t}}^2 + }
	{}{}
	\breakObjective{\lagrangemultiplier\T (\embedsensing_{t+1} - \embedsensing_{t} - (\linearlatentdynA_{t} \embedsensing_{t} + \linearlatentdynB_{t} \action_{t} + \linearlatentdync_{t}) \deltat)}
\end{mini}
Setting the partial derivative of:
\begin{align*}
    L_{CO}(\action_{t}, \embedsensing_{t+1}, \lagrangemultiplier) = &\frac{1}{2} \norm{\embedsensing_{T} - \embedsensing_{t+1}}^2 + \frac{\beta}{2} \norm{\action_{t}}^2 + \\
    &\lagrangemultiplier\T (\embedsensing_{t+1} - \embedsensing_{t} - (\linearlatentdynA_{t} \embedsensing_{t} + \linearlatentdynB_{t} \action_{t} + \linearlatentdync_{t}) \deltat) \numberthis
\end{align*}
w.r.t. the optimization variables to zero, gives us:
\begin{equation}
    \frac{\partial L_{CO}}{\partial \lagrangemultiplier} = \mathbf{0} \Rightarrow \embedsensing_{t+1} = \embedsensing_{t} + (\linearlatentdynA_{t} \embedsensing_{t} + \linearlatentdynB_{t} \action_{t} + \linearlatentdync_{t}) \deltat
    \label{eq:partial_L_partial_lagrangemultiplier}
\end{equation}
\begin{equation}
    \frac{\partial L_{CO}}{\partial \action_{t}} = \mathbf{0} \Rightarrow \beta \action_{t} - \linearlatentdynB_{t}\T \lagrangemultiplier \deltat = \mathbf{0} \Rightarrow \action_{t} =  \frac{\deltat}{\beta} \linearlatentdynB_{t}\T \lagrangemultiplier
    \label{eq:partial_L_partial_action_t}
\end{equation}
\begin{equation}
    \frac{\partial L_{CO}}{\partial \embedsensing_{t+1}} = \mathbf{0} \Rightarrow -(\embedsensing_{T} - \embedsensing_{t+1}) + \lagrangemultiplier = \mathbf{0} \Rightarrow \embedsensing_{t+1} = \embedsensing_{T} - \lagrangemultiplier
    \label{eq:partial_L_partial_embedsensing_tplus1}
\end{equation}
Combining Eq. \ref{eq:partial_L_partial_lagrangemultiplier}, \ref{eq:partial_L_partial_embedsensing_tplus1}, and \ref{eq:partial_L_partial_action_t}:
\begin{align}
    \embedsensing_{T} - \lagrangemultiplier &= \embedsensing_{t} + (\linearlatentdynA_{t} \embedsensing_{t} + \linearlatentdynB_{t} \action_{t} + \linearlatentdync_{t}) \deltat \nonumber \\
    \embedsensing_{T} - \lagrangemultiplier &= \embedsensing_{t} + \linearlatentdynA_{t} \embedsensing_{t} \deltat + \frac{{\deltat}^2}{\beta} \linearlatentdynB_{t} \linearlatentdynB_{t}\T \lagrangemultiplier + \linearlatentdync_{t} \deltat \\
    \left(\frac{{\deltat}^2}{\beta} \linearlatentdynB_{t} \linearlatentdynB_{t}\T + \eye\right) \lagrangemultiplier &= \embedsensing_{T} - \embedsensing_{t} - \linearlatentdynA_{t} \embedsensing_{t} \deltat - \linearlatentdync_{t} \deltat \nonumber
\end{align}
\begin{equation}
    \lagrangemultiplier = \frac{\beta}{\deltat} \left(\linearlatentdynB_{t} \linearlatentdynB_{t}\T + \frac{\beta}{{\deltat}^2} \eye\right)^{-1} \left(\frac{\embedsensing_{T} - \embedsensing_{t}}{\deltat} - \linearlatentdynA_{t} \embedsensing_{t} - \linearlatentdync_{t} \right)
    \label{eq:lagrangemultiplier_solution}
\end{equation}
Finally we substitute Eq. \ref{eq:lagrangemultiplier_solution} into Eq. \ref{eq:partial_L_partial_action_t}, we get:
\begin{equation}
    \action_{t} = \linearlatentdynB_{t}\T \left(\linearlatentdynB_{t} \linearlatentdynB_{t}\T + \frac{\beta}{{\deltat}^2} \eye\right)^{-1} \left(\frac{\embedsensing_{T} - \embedsensing_{t}}{\deltat} - \linearlatentdynA_{t} \embedsensing_{t} - \linearlatentdync_{t}\right)
\end{equation}

\section*{ACKNOWLEDGMENT}
We thank David Crombecque and Ragesh Kumar Ramachandran, both from the University of Southern California, for the insightful discussions on mathematical manifolds. We also thank Arunkumar Byravan for discussions on the SE3-Pose-Nets paper, and Kendall Lowrey for the help on a finishing work of the BioTac mounting for an earlier version of the work, both from the University of Washington.


\bibliographystyle{IEEEtran}
\bibliography{references}

\end{document}